\documentclass[letterpaper]{article}
\usepackage{uai2020}
\usepackage[margin=1in]{geometry}

\usepackage{times}
\usepackage[dvipsnames]{xcolor}
\usepackage[colorlinks,allcolors=BlueViolet,bookmarks=false,hypertexnames=true]{hyperref}
\usepackage{graphicx}
\usepackage{amsmath,amsfonts}
\usepackage{algorithm}
\usepackage{algorithmic,amssymb,xspace,nicefrac,etoolbox}
\usepackage{microtype}
\usepackage{array}
\usepackage{balance}

\usepackage{listings}
\usepackage{lstautogobble}  
\usepackage{color}          
\usepackage{zi4}            
\definecolor{bluekeywords}{rgb}{0.13, 0.13, 1}
\definecolor{greencomments}{rgb}{0, 0.5, 0}
\definecolor{redstrings}{rgb}{0.9, 0, 0}
\definecolor{graynumbers}{rgb}{0.5, 0.5, 0.5}
\usepackage{anyfontsize}
\usepackage[round]{natbib}
\title{Locally Masked Convolution for Autoregressive Models}

\author{ {\bf Ajay Jain} \\
UC Berkeley \\
ajayj@berkeley.edu \\
\And
{\bf Pieter Abbeel} \\
UC Berkeley \\
pabbeel@berkeley.edu \\
\And
{\bf Deepak Pathak} \\
Carnegie Mellon University \\
dpathak@cs.cmu.edu \\
}

\begin{document}

\newcommand{\todo}[1]{{\color{red} TODO: #1}}
\newcommand{\ie}[0]{\textit{i.e.}}
\newcommand{\eg}[0]{\textit{e.g.}}
\newcommand{\ours}[0]{\textsc{LMConv}}

\maketitle

\begin{abstract}
High-dimensional generative models have many applications including image compression, multimedia generation, anomaly detection and data completion.
State-of-the-art estimators for natural images are autoregressive, decomposing the joint distribution over pixels into a product of conditionals parameterized by a deep neural network, \eg{} a convolutional neural network such as the PixelCNN. However, PixelCNNs only model a single decomposition of the joint, and only a single generation order is efficient. For tasks such as image completion, these models are unable to use much of the observed context. To generate data in arbitrary orders, we introduce \textsc{LMConv}: a simple modification to the standard 2D convolution that allows arbitrary masks to be applied to the weights at each location in the image. Using \textsc{LMConv}, we learn an ensemble of distribution estimators that share parameters but differ in generation order, achieving improved performance on whole-image density estimation (2.89 bpd on unconditional CIFAR10), as well as globally coherent image completions. Our code is available at {\small \url{https://ajayjain.github.io/lmconv}}.
\end{abstract}

\begin{figure}[!t]
	\centering
    \includegraphics[width=0.8\linewidth]{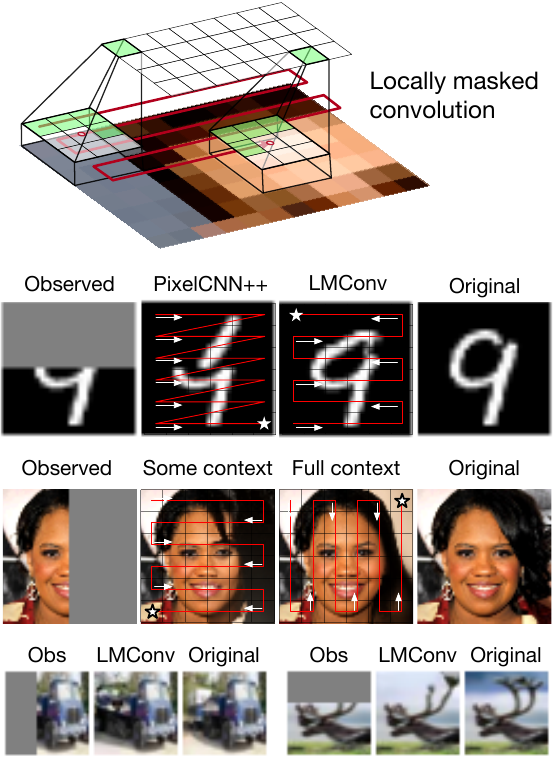}
    \vspace{-3mm}
	\caption{The ideal autoregressive joint distribution decomposition and sampling order are task-dependent. We learn to generate images under multiple orderings with the same parameters via \textit{locally masked convolutions} (top), enabling global coherence for image completion (bottom).}
	\label{fig:inpainting}
	\vspace{-2mm}
\end{figure}

\section{INTRODUCTION}
Learning generative models of high-dimensional data such as images is a holy grail of machine learning with pervasive applications. 
Significant progress on this problem would naturally lead to a wide range of applications, including multimedia generation, compression, probabilistic time series forecasting, representation learning, and missing data completion.
Many generative modeling frameworks have been proposed.
Current state-of-the-art models for high-dimensional image data include
(a) autoregressive models~\citep{bengio2000modeling,efros1999texture}, (b) normalizing flow density estimators~\citep{rezende2015variational}, (c) generative adversarial networks (GANs) \citep{goodfellow2014generative}, (d) latent variable models such as the VAE~\citep{kingma2013auto,rezende2014stochastic} and (e) energy-based models \eg{}~\cite{hinton2002training, lecun2006tutorial, NIPS2019_8619, song2019generative}. While GANs, VAEs and EBMs have had great success in high-dimensional image generation, exact likelihoods are generally intractable. Likelihood estimation is key for many practical applications from uncertainty estimation, robustness, reliability and safety perspectives.
In contrast, autoregressive and flow models estimate exact likelihoods and can be used for uncertainty estimation, though still have room for improved generation quality. In this work, our focus is on autoregressive models.

\begin{figure}[t]
    \centering
    \includegraphics[width=0.8\linewidth]{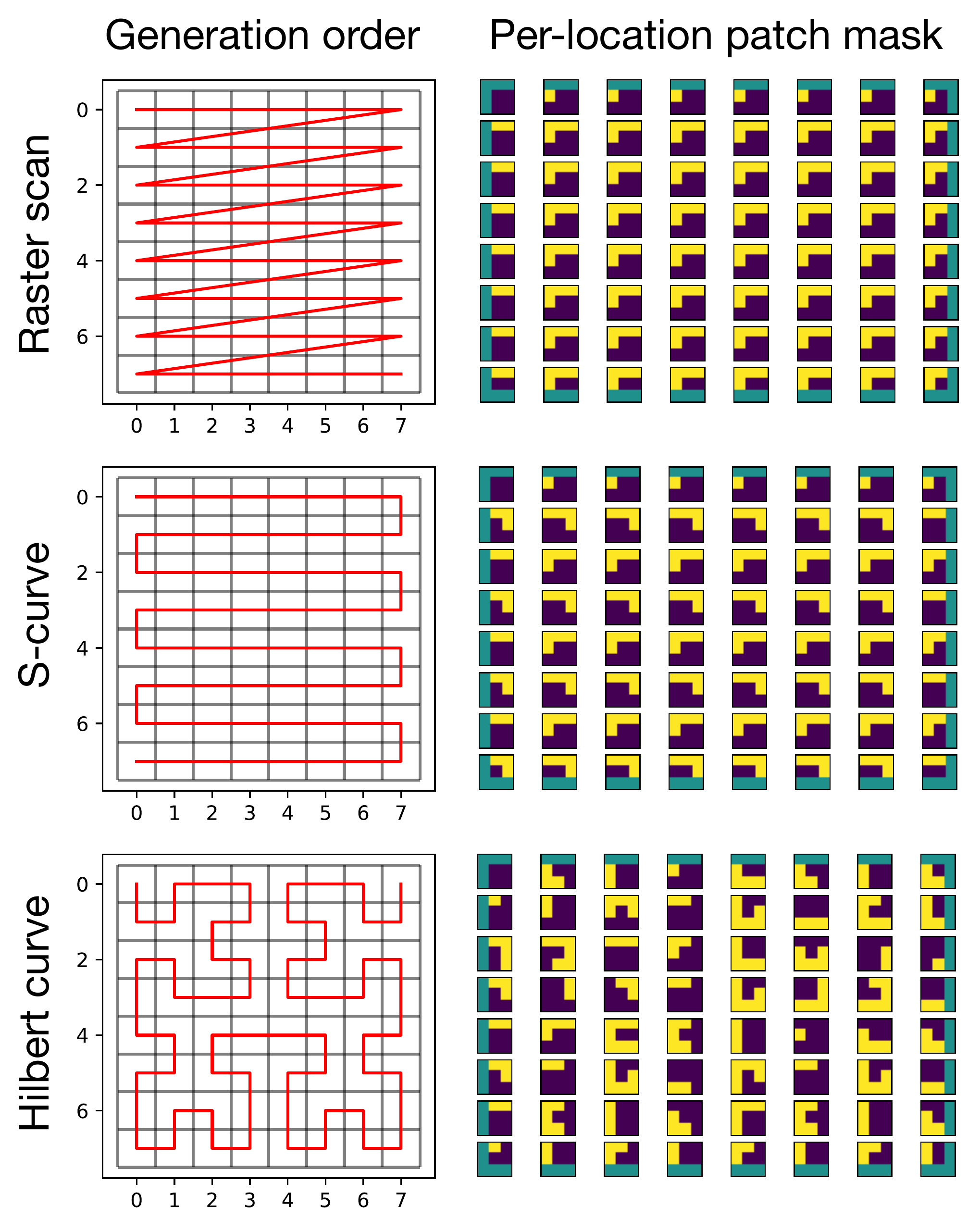}
    \vspace{-4mm}
    \caption{The three pixel generation orders and corresponding local masks that we consider in this work.}
    \label{fig:orders_viz}
\end{figure}

Given $n$ variables, one can generate $n!$ autoregressive decompositions of the joint likelihood, each corresponding to a forward sampling order, and more if we assume conditional independence. Early autoregressive texture synthesis~\citep{popat1993novel,efros1999texture} work could support multiple orders. However, recent CNN-based autoregressive models for images~\citep{oord2016pixel,van2016conditional,salimans2017pixelcnnpp} capture only one of these orders (typically left-to-right raster scan, Fig. \ref{fig:orders_viz}) for practical computational efficiency. 
Training and testing with a single order will not support all scenarios. Consider the image completion task in first row of Figure~\ref{fig:inpainting}. If the top half of the image is missing, a raster scan generation order from left-to-right and top-to-bottom does not allow the model to condition on the context given in the observed bottom half of the image as the required conditionals are not estimated by the model.

In this work, we propose a scalable, yet simple modification to convolutional autoregressive models to estimate more accurate likelihoods with a minor change in computation during training. 
Our goal is to support arbitrary orders in a scalable manner, allowing more precise likelihoods by averaging over several graphical models corresponding to orders (a form of Bayesian model averaging). Some past works have supported arbitrary orders in autoregressive models by learning separate parameters for each model~\citep{frey1998graphical}, or by masking the input image to hide successor variables~\citep{larochelle2011neural}. A more efficient approach is to estimate densities in parallel across dimensions by masking network weights~\citep{germain2015made} differently for each order. However, all these methods are still computationally inefficient and difficult to scale beyond fully-connected networks to convolutional architectures.

In this work, we perform order-agnostic distribution estimation for natural images with state-of-the-art convolutional architectures.
We propose to support arbitrary orderings by introducing masking at the level of features, rather than on inputs or weights. 
We show how an autoregressive CNN can support and learn multiple orders, with a single set of weights, via \textit{locally masked convolutions} that efficiently apply location-specific masks to patches of each feature map. 
These local convolutions can be efficiently implemented purely via matrix multiplication by incorporating masking at the level of the im2col and col2im separation of convolution~\citep{jia2014caffe}. 

Arbitrary orders allow us to customize the traversal based on the needs of the task, which we evaluate in experiments. For instance, consider the examples shown in Fig.~\ref{fig:inpainting}. The flexibility allows us to select the sampling order that exposes the maximum possible context for image completion, choose orderings that eliminate blind-spots (unobservable pixels) in image generation, and ensemble across multiple orderings using the same network weights. Note that such a model is able to support these image completions without training on any inpainting masks.

In experiments, we show that our approach can be efficiently implemented and is flexible without sacrificing the overall distribution estimation performance. By introducing order-agnostic training via \ours{}, we significantly outperform PixelCNN++ on the unconditional CIFAR10 dataset, achieving code lengths of 2.89 bits per dimension. We show that the model can generalize to some novel orders. Finally, we significantly outperform raster-scan baselines on conditional likelihoods relevant to image completion by customizing the generation order.

\begin{figure*}[!t]
	\centering
	\includegraphics[width=\linewidth]{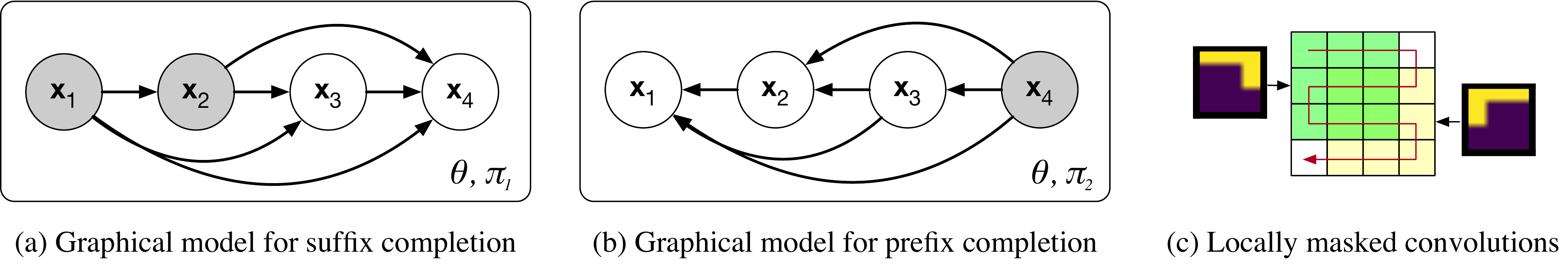}
	\vspace{-8mm}
	\caption{(a) A graphical model where the final, unobserved variables $x_3, x_4$ can be efficiently completed via forward sampling conditioned on the observed variables $x_1, x_2$. (b) When $x_4$ is observed, we sample $x_1, x_2,$ and $x_3$ in the second graphical model using the same parameters. (c) \ours{} defines the model with masks at each filter location.}
	\label{fig:method}
\end{figure*}

\section{BACKGROUND}
Deep autoregressive models estimate high-dimensional data distributions using samples from the joint distribution over D-dimensions $p_\text{data}(\mathbf{x}_1, \ldots, \mathbf{x}_D)$. In this setting, we wish to approximate the joint with a parametric model $p_\mathbf{\theta}(\mathbf{x}_1, \ldots, \mathbf{x}_D)$ by minimizing KL-divergence $D_{KL}(p_\text{data} || p_\mathbf{\theta})$, or equivalently by maximizing the log-likelihood of the samples. As a general modeling principle, we can divide high-dimensional variables into many low-dimensional parts such as single dimensions, and capture dependencies between dimensions with a directed graphical model. Following the notation of \citep{kingma2019introduction}, these autoregressive (AR) models represent the joint distribution as a product of conditionals,
\begin{align}
	p_\mathbf{\theta}(\mathbf{x}) &= p_\mathbf{\theta}(x_1, \ldots, x_D) \nonumber\\
	&= p_\mathbf{\theta}(x_{\pi(1)}) \prod_{i=2}^D p_\mathbf{\theta}\left(x_{\pi(i)} \mid Pa(\mathbf{x}_{\pi(i)})\right) \label{eq:autoreg_factorization}
\end{align}
where $\pi : [D] \rightarrow [D]$ is a permutation defining an order over the dimensions, $Pa(\mathbf{x}_{\pi(i)}) = \mathbf{x}_{\pi(1)}, \ldots, \mathbf{x}_{\pi(i-1)}$ defines the parents of $x_{\pi(i)}$ in the graphical model, and $\theta$ is a parameter vector. 
As any joint can be decomposed in this manner according to the product rule, this factorization provides the foundation for many models including ours. The primary challenge in autoregressive models is defining a sufficiently expressive family for the conditionals where parameter estimation is efficient. Deep autoregressive models parameterize the conditionals with a neural network that is provided the context $Pa(\mathbf{x}_{\pi(i)})$.

Decomposition \eqref{eq:autoreg_factorization} converts the joint modeling problem into a sequence modeling problem. Forward (ancestral) sampling draws root variable $x_{\pi(1)}$ first, then samples the remaining dimensions in order $x_{\pi(2)}, \ldots, x_{\pi(D)}$ from their respective conditionals. Given a particular autoregressive decomposition of the joint, forward sampling supports a single data generation order. The joint model density for an observed variable can be computed exactly by evaluating each conditional, allowing density estimation and maximum likelihood parameter estimation,
\begin{align}
\mathcal{L}(\theta) &= \mathbb{E}_{x\sim p_\text{data}} \sum_{i=1}^D \log p_\mathbf{\theta}\left(x_{\pi(i)} \mid x_{\pi(1)}, \ldots, x_{\pi(i-1)} \right) \nonumber\\
\theta^* &= \arg_\mathbf{\theta}\max \mathcal{L}(\theta)
\end{align}

With some choices of network architecture, the conditionals can be computed in parallel by masking weights \citep{germain2015made, oord2016pixel}. In the PixelCNN model family, masked convolutions are \textit{causal}: the features output by a masked convolution can only depend on features earlier in the order.

While the choice of order is arbitrary, temporal and sequential data modalities have a natural ordering from the first dimension in the sequence to the last. For spatial data such as images, a natural ordering is not clear. For computational reasons, a \textit{raster scan} order is generally used where the top left pixel is modeled unconditionally and generation proceeds in row-major fashion across each row from left to right, depicted in Figure~\ref{fig:inpainting}, second column.

\section{IMAGE COMPLETION WITH MAXIMUM RECEPTIVE FIELD}

For estimating the distribution of 2D images, a raster scan ordering is perhaps as good of an order as any other choice. That said, the raster scan order has necessitated architectural innovations to allow the neural network to access information far back in the sequence such as two-dimensional PixelRNNs \citep{oord2016pixel}, two-stream shift-based convolutional architectures \citep{van2016conditional}, and self-attention combined with convolution \citep{chen2017pixelsnail}. These structures significantly improve test-set likelihoods and sample quality, but marry network architectures to the raster scan order.

Fixing a particular order is limiting for missing data completion tasks. Letting $\pi(i) = i$ denote the raster scan order, PixelRNN and PixelCNN architectures can complete only the bottom part of the image via forward sampling: given observations $x_1,\ldots,x_d$, raster scan autoregressive models sequentially sample, \begin{equation}
\hat{x}_i \sim p_\mathbf{\theta}(x_i \mid x_1,\ldots,x_d, \hat{x}_{d+1}, \ldots, \hat{x}_{i-1}).
\end{equation}
If all dimensions other than $x_i$ are observed, ideally we would sample $\hat{x}_i$ using maximum conditioning context, \begin{equation}
\hat{x}_i \sim p_\mathbf{\theta}(x_i \mid x_{<i}, x_{> i}). \label{eq:ideal}   
\end{equation}
Unfortunately, the raster scan model only predicts distributions of the form $p_\mathbf{\theta}(x_i \mid x_{<i})$, and ignores observations $x_{> i}$ during completion.
In the worst case, a model with a raster scan generation order \textit{cannot observe any of the context} for an inpainting task where the top half of the image is unknown ({Figure~\ref{fig:inpainting}, PixelCNN++}). This leads to image completions that do not respect global structure.
Small numbers of dimensions could be sampled by computing the posterior, \eg{} for $i=1$,
\begin{equation}
	p_\theta(\hat{x}_1 \mid x_{>1}) = \frac{p_\theta(\hat{x}_1, x_{>1})}{\sum_{x_1'} p_\theta(x_1', x_{>1})},
\end{equation}
but this is expensive as each summand requires neural network evaluation, and becomes intractable when several dimensions are unknown. 
Instead of approximating the posterior, we estimate parameters $\mathbf{\theta}$ that achieve high likelihood with multiple autoregressive decompositions,
\begin{align}
\mathcal{L}_\text{OA}(\theta) &= \mathbb{E}_{x\sim p_\text{data}} \mathbb{E}_{\pi \sim p_{\pi}} \log p_\mathbf{\theta}\left(x_1, \ldots, x_D ; \pi \right) \nonumber\\
\theta^* &= \arg_\mathbf{\theta}\max \mathcal{L}_\text{OA}(\theta) \label{eq:optimization}
\end{align}
with $p_\pi$ denoting a uniform distribution over several orderings. The joint distribution under $\pi$ factorizes according to \eqref{eq:autoreg_factorization}. The resulting conditionals are all parameterized by the same neural network. By choosing order prior $p_\pi$ that supports a $\pi$ such that $\pi(D) = i$, we can use the network with such an ordering to query \eqref{eq:ideal} directly.

During optimization with stochastic gradient descent, we make single-sample estimates of the inner expectation in \eqref{eq:optimization} according to order-agnostic training \citep{uria2014deep, germain2015made}, using a single order per batch.

For a test-time task where $\{ x_i : i \in T_\text{obs}\}$ are observed, we select a $\pi$ that the model was trained with such that $$\{\pi(1),\ldots,\pi(|T_\text{obs}|)\} = T_\text{obs},$$ \ie{} the first $|T_\text{obs}|$ dimensions in the generation order are the observed dimensions, then sample according to the rest of the order so that the model posterior over each unknown dimension is conditioned either on observed or previously sampled dimensions.

\section{LOCAL MASKING}

In this section, we develop \textit{locally masked convolutions} (\ours{}): a modification to the standard convolution operator that allows control over generation order and parallel computation of conditionals for evaluating likelihood.
In the first convolutional layer of a neural network, $C_\text{out}$ filters of size $k \times k$ are applied to the input image with spatial invariance: the same parameters are used at all locations in a sliding window. Each filter has $k^2 * C_\text{in}$ parameters. For images with discretized intensities, convolutional autoregressive networks transform a spatial $H \times W$, multi-channel image into a tensor of log-probabilities that define the conditional distributions of \eqref{eq:autoreg_factorization}. These log-probabilities take the form of an $H \times W$ image, with channel count equal to the number of color channels times the number of bins per color channel. The output log-probabilities at coordinate $i, j$ in the output define the distribution $p_\theta(x_{i, j} \mid Pa(p(x_{i,j}))$. Critically, this distribution must not depend on observations of successors in the Bayesian network, or the product of conditionals will not define a valid distribution due to cyclicity.

\begin{figure}
    \centering
    \includegraphics[width=\linewidth]{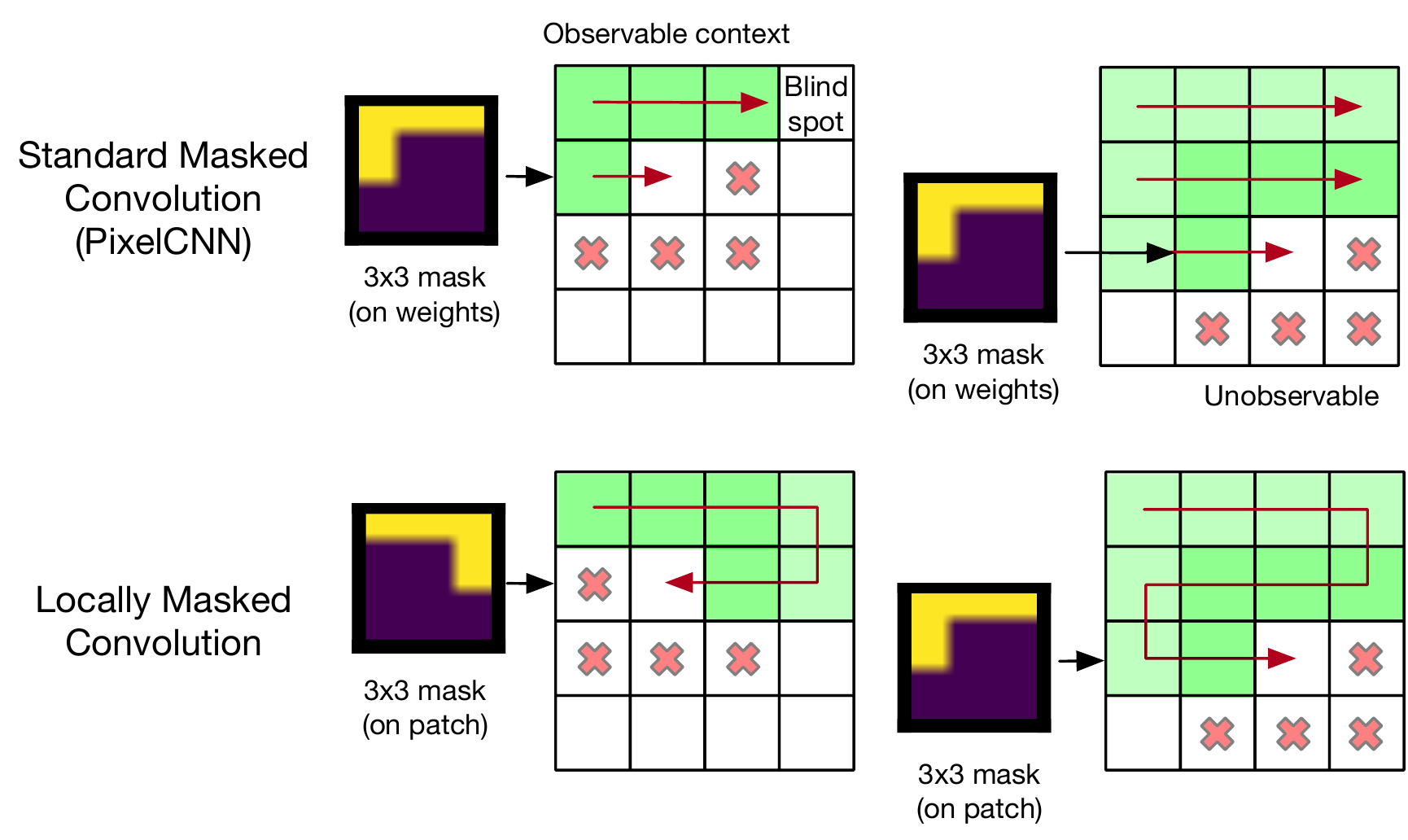}
    \vspace{-8mm}
    \caption{A comparison of standard weight masked convolutions and the proposed locally masked convolution.}
    \label{fig:convs}
    \vspace{-1mm}
\end{figure}

NADE~\citep{larochelle2011neural} circumvents the problem by masking the input image, though requires independent forward passes to compute each factor of the autoregressive decomposition \eqref{eq:autoreg_factorization}. Instead, the PixelCNN model family controls information flow through the network by setting certain weights of the convolution filters to zero, similar to how MADE \citep{germain2015made} masks the weight matrices in fully-connected layers. We depict masked convolutions for the first convolutional layer in Figure~\ref{fig:convs}. As a single mask is applied to the $C_\text{in} \times k \times k$ parameter tensor defining each convolutional filter, the same masking pattern is applied at all locations in the image. Sharing the masking pattern constrains the possible orders, and leads to blind spots which the output distribution is unable to observe.

In practice, convolutions are implemented through general matrix multiplication (GEMM) due to widely available, heavily optimized and parallelized implementations of the operation on GPU and CPU. To use matrix multiplication, the input to a layer is rearranged in memory via the im2col algorithm, which extracts $C_\text{in} \times k \times k$ patches from the $C_\text{in} \times H \times W$ input at each location that a convolutional filter will be applied. Assuming padding and a stride of $1$ is used, the rearrangement yields matrix $X$ with $C_\text{in} * k^2$ rows and $H*W$ columns. To perform convolution, the framework left-multiplies weight matrix $\mathcal{W}$, storing $Y = \mathcal{W}X$, adds a bias, and finally rearranges $Y$ into a spatial format via the col2im algorithm.

We exploit this data rearrangement to arbitrarily mask the input to the convolutional filter at each location it is applied. The inputs to the convolution at each location, \ie{} the input patches, form columns of $X$. For a given generation order, we construct mask matrix $\mathcal{M}$ of the same dimensions as $X$ and set $X = \mathcal{M} \odot X$ prior to matrix multiplication. In particular, our locally masked convolution masks \textit{patches of the input to each layer}, rather than masking weights and rather than masking the initial input to the network. \textsc{LMConv} combines the flexibility of NADE and the parallelizability of MADE and PixelCNN. 
The \textsc{LMConv} algorithm is summarized in Algorithm~\ref{alg:lmconv}, and mask construction is detailed in Algorithm~\ref{alg:makemasks}.

We implement two versions of the layer with the PyTorch machine learning framework~\citep{NEURIPS2019_9015}. The first is an implementation that uses autodifferentiation to compute gradients. As only the forward pass is defined by the user, the implementation is under 20 lines of Python.

However, reverse-mode autodifferentiation incurs significant memory overheads during backpropagation as the output of nearly every operation during the forward pass must be stored until gradient computation \citep{griewank2000algorithm, mlsys2020_196}. Data rearrangement with im2col is memory intensive as features patches overlap and are duplicated. We implement a custom, memory efficient backward pass that only stores the input, the mask and the output of the layer during the forward pass and recomputes the im2col operation during the backward pass. Recomputing the im2col operation achieves 2.7$\times$ memory savings at a $1.3\times$ slowdown.

Using locally masked convolutions, we can experiment with many different image generation orders. In this work, we consider three classes of orderings: raster scan, implemented in baseline PixelCNNs, an S-curve order that traverses rows in alternating directions, and a Hilbert space-filling curve order that generates nearby pixels in the image consecutively. Alternate orderings provide several benefits. Nearby pixels in an image are highly correlated. By generating these pixels close in a Hilbert curve order, we might expect information to propagate from the most important, nearby observations for each dimension and reduce the vanishing gradient problem.

\begin{algorithm}[t]
   \caption{\textsc{LMConv}: Locally masked 2D convolution}
   \label{alg:lmconv}
\begin{algorithmic}[1]
   \STATE {\bfseries Input:} image $x$, weights $\mathcal{W}$, bias $b$, generation order $\pi$. $x$ is ${B \times C_\text{in} \times H \times W}$ dimensional and $\mathcal{W}$ is ${C_\text{out} \times C_\text{in} * k_1 * k_2}$ dimensional
   \STATE Create mask matrix $\mathcal{M}$ with Algorithm~\ref{alg:makemasks}
   \STATE Extract patches: $X = \texttt{im2col}(\texttt{pad}(x), k_1, k_2)$
   \STATE Mask patches: $X = \mathcal{M} \odot X$
   \STATE Perform convolution via batch MM: $Y = \mathcal{W}X + b$
   \STATE Assemble patches: $y = \texttt{col2im}(Y)$
   \STATE {\bfseries return} $y$
\end{algorithmic}
\end{algorithm}

If the image is considered a graph with nodes for each pixel and edges connecting adjacent pixels, a convolutional autoregressive model using an order defined by a Hamiltonian path over the image graph will also suffer no blind spot in a $D$ layer network. To see this, note that the features corresponding to dimension $x_{\pi(i)}$ in the Hamiltonian path order will always be able to observe the previous layer's features corresponding to $x_{\pi(i-1)}$. After at least $D$ layers of depth, the features for $x_{\pi(i)}$ will incorporate information from all $i-1$ previous dimensions. In practice, information propagates with fewer required layers in these architectures as multiple neighbors are observed in each layer. Finally, we select multiple orderings at inference and average the resulting joint distributions to compute better likelihood estimates.

\begin{algorithm}[t]
   \caption{Create input mask matrix}
   \label{alg:makemasks}
\begin{algorithmic}[1]
   \STATE {\bfseries Input:} Generation order $\pi(\cdot)$, constants $C_\text{in}, k_1, k_2$, dilation $d$, is this the first layer?
   \STATE Start with an empty set of generated coordinates
   \STATE Initialize $\mathcal{M}$ as $k_1 * k_2 \times H*W$ zero matrix
   \FOR{$i$ from $1$ to $H*W$}
   		\STATE Let $(r, c)$ be coordinates of dimension $\pi(i)$
   		\FOR{offsets $\Delta_r, \Delta_c$ in $k_1 \times k_2$ kernel}
	   		\IF{$(r + d\Delta_r, c + d\Delta_c)$ has been generated}
	   			\STATE Allow output location $(r, c)$ to access features at $(r + d\Delta_r, c + d\Delta_c)$ in previous layer: set $\mathcal{M}_{k_2\Delta_r+\Delta_c, Wr+c} = 1$
	   		\ENDIF
   		\ENDFOR{}
   		\STATE Add $(r, c)$ to generated coordinates
   \ENDFOR{}
   \IF{not the first layer}
   		\STATE Allow previous layer features to be observed at all locations: set center row $\lfloor \frac{k_1 * k_2 }{2} \rfloor$ of $\mathcal{M}$ to 1
   \ENDIF
   \STATE Repeat rows of $\mathcal{M}$, $C_\text{in}$ times
   \STATE {\bfseries return} binary mask matrix $\mathcal{M}$
\end{algorithmic}
\end{algorithm}

\section{ARCHITECTURE}
\label{sec:architecture}

We use a network architecture similar to PixelCNN++ \citep{salimans2017pixelcnnpp}, the best-in-class density estimator in the fully convolutional autoregressive PixelCNN model family. Convolution operations are masked according to Algorithm~\ref{alg:lmconv}. While our locally masked convolutions can benefit from self-attention mechanisms used in later work, we choose a fully convolutional architecture for simplicity and to study the benefit of local masking in isolation of other architectural innovations. We make three modifications to the PixelCNN++ architecture that simplify it and allow for arbitrary generation orders.
Gated PixelCNN uses a two-stream architecture composed of two network stacks with $\lfloor \frac{k}{2} \rfloor \times 1$ and $\lfloor \frac{k}{2} \rfloor \times k$ 
convolutions to enforce the raster scan order. In the horizontal stream, Gated PixelCNN applies non-square convolutions and feature map shifts or pads to extract information within the same row, to the left of the current dimension. In the vertical stream, Gated PixelCNN extracts information from above. Skip connections between streams allow information to propagate.
PixelCNN$++$ uses a similar architecture based on a U-Net~\citep{ronneberger2015u} with approximately 54M parameters. We replace the two streams with a simple, single stream with the same depth, using \ours{} to maintain the autoregressive property. Masks for these convolutions are computed and cached at the beginning of training. Due to the regularizing effect of order-agnostic training, we do not use dropout.

\begin{table}[t]
\caption{Average negative log likelihood of binarized and grayscale MNIST digits under our model. Lower is better.}
\label{table:bpd_mnist}
\begin{center}
\resizebox{\linewidth}{!}{\begin{tabular}{lc}
\multicolumn{1}{l}{\bf BINARIZED MNIST, 28x28}  &\multicolumn{1}{c}{\bf NLL (nats)} \\
\hline \\
DARN (Intractable) \citep{gregor2014deep} & $\approx$84.13 \\
NADE \citep{uria2014deep}		& 88.33 \\
EoNADE 2hl (128 orders) \citep{uria2014deep} & 85.10 \\
EoNADE-5 2hl (128 orders) \citep{raiko2014iterative} & 84.68 \\
MADE 2hl (32 orders) \cite{germain2015made}	 		& 86.64 \\
PixelCNN \citep{oord2016pixel}		& 81.30 \\
PixelRNN \citep{oord2016pixel}		& 79.20 \\
Ours, S-curve (1 order)  & 78.47 \\
Ours, S-curve (8 orders)  &  \textbf{77.58} \\\\
\multicolumn{1}{l}{\bf GRAYSCALE MNIST, 28x28}  &\multicolumn{1}{c}{\bf NLL (bpd)} \\
\hline \\
Spatial PixelCNN \citep{akoury2017spatial}    & 0.88              \\
PixelCNN++ (1 stream)    & 0.77              \\
Ours, S-curve (1 order) & 0.68 \\
Ours, S-curve (8 orders)      &  \textbf{0.65} \\
\end{tabular}}
\end{center}
\vspace{-2mm}
\end{table}

Second, we use dilated convolutions \citep{yu2015multi} at regular intervals in the model rather than downsampling the feature map. Downsampling precludes many orders, as the operation aggregates information from contiguous squares of pixels together without a mask. Dilated convolutions expand the receptive field without limiting the order, as local masks can be customized to hide or reveal specific features accessed by the filter.

Finally, we normalize the feature map across the channel dimension~\citep{li2019positional}. Normalization allows masks to have varying numbers of ones at each spatial location by rescaling features to the same scale.

As in PixelCNN++, our model represents each conditional with a mixture of 10 discretized logistic distributions that imposes a distribution over binned pixel intensities. For the binarized MNIST dataset \citep{salakhutdinov2008quantitative}, we instead use a softmax over two logits. We train with 8 variants of an S-curve (zig-zag) order that traverses each row of the image in alternating directions so that consecutively generated pixels are adjacent, and so that locally masked CNNs with sufficient depth can achieve the maximum allowed receptive field.

Across all quantitative experiments, we use a model with approximately 46\textsc{M} parameters, trained with the Adam optimizer with a learning rate of $2*10^{-4}$ decayed by a factor of $1-5*10^{-6}$ per iteration with clipped gradients. For CelebA-HQ qualitative results, we increase filter count and train a model with 184\textsc{M} parameters. More details are provided in the appendix.

\begin{table}[t]
\caption{Average negative log likelihood of CIFAR10 images under our model. Lower is better.}
\label{table:bpd_cifar}
\begin{center}
\resizebox{\linewidth}{!}{\begin{tabular}{lc}
\multicolumn{1}{l}{\bf CIFAR10, 32x32}  &\multicolumn{1}{c}{\bf NLL (bpd)} \\
\hline \\
Uniform Distribution & 8.00 \\
Multivariate Gaussian \citep{oord2016pixel} & 4.70 \\
\textbf{Attention-based} & \\
Image Transformer \citep{parmar2018image}  & 2.90 \\
PixelSNAIL \citep{chen2017pixelsnail}        & 2.85 \\
Sparse Transformer \citep{child2019generating}  & \textbf{2.80} \\
\textbf{Convolutional} & \\
PixelCNN (1 stream) \citep{oord2016pixel}	& 3.14\\
Gated PixelCNN (2 stream) \citep{van2016conditional} &	3.03 \\
PixelCNN++ (1 stream) & 2.99 \\
PixelCNN++ (2 stream) \citep{salimans2017pixelcnnpp}   & 2.92    \\
Ours, S-curve (1 stream, 1 order) & 2.91 \\
Ours, S-curve (1 stream, 8 orders) & \textbf{2.89}
\end{tabular}}
\end{center}
\vspace{-2mm}
\end{table}

\section{EXPERIMENTS}

To evaluate the benefits of our approach, we study three scientific questions: (1) \textit{do locally masked autoregressive ensembles estimate more accurate likelihoods on image datasets than single-order models?}, (2) \textit{can the model generalize to novel orders?} and (3) \textit{how important is order selection for image completion?}

We estimate the distribution of three image datasets: 28$\times{}$28 grayscale and binary \citep{salakhutdinov2008quantitative} MNIST digits, 32$\times{}$32 8-bit color CIFAR10 natural images, and high-resolution CelebA-HQ 5-bit color face photographs \citep{karras2018progressive}. Unlike classification, density estimation remains challenging on these datasets. We train the CelebA-HQ models at 256$\times{}$256 resolution to compare with prior density estimation work, and at a bilinearly downsampled 64$\times$64 resolution.

Our locally masked model achieves better likelihoods than PixelCNN++ by using multiple generation orders. We then show that the model can generalize to generation orders that it has not been trained with. Finally, for image completion, we achieve the best results over strong baselines by using orders that expose all observed pixels.

\subsection{WHOLE-IMAGE DENSITY ESTIMATION}

Tractable generative models are generally evaluated via the average negative log likelihood (NLL) of test data. For interpretability, many papers normalize base 2 NLL by the number of dimensions. By normalizing, we can measure bits per dimension (bpd), or a lower-bound for the expected number of bits needed per pixel to losslessly compress images using a Huffman code with $p(\mathbf{x})$ estimated by our model. Better estimates of the distribution should result in higher compression rates. Tables~\ref{table:bpd_mnist} and \ref{table:bpd_cifar} show likelihoods for our model and prior models.

On binarized MNIST (Table~\ref{table:bpd_mnist}), our locally masked PixelCNN achieves significantly higher likelihoods (lower NLL) than baselines, including neural autoregressive models NADE, EoNADE, and MADE that average across large numbers of orderings. This is due to architectural advantages of our CNN and increased model capacity. Our model also outperforms the standard PixelCNN, which suffers from a blind spot problem due to sharing the same mask at all locations. Likelihood is further improved by using ensemble averaging across 8 orders that share parameters. These results are also observed on grayscale MNIST where each pixel has one of 256 intensity levels.

\begin{table}
\caption{Average conditional negative log likelihood for \textbf{T}op, \textbf{L}eft and \textbf{B}ottom half image completion.}
\vspace{-2mm}
\label{table:mnistbpd}
\begin{center}
\resizebox{\linewidth}{!}{\begin{tabular}{lccc}
\multicolumn{1}{l}{\bf BINARIZED MNIST 28x28 (nats)}  &\multicolumn{1}{c}{\bf T} &\multicolumn{1}{c}{\bf L} &\multicolumn{1}{c}{\bf B} \\
\hline \\
Ours (adversarial order) & 41.76 & 39.83 & 43.35 \\
Ours (1 max context order) & 34.99 & 32.47 & 36.57 \\
Ours (2 max context orders) & \textbf{34.82} & \textbf{32.25} & \textbf{36.36} \\\\
\multicolumn{1}{l}{\bf CIFAR10 32x32 (bpd)}  &\multicolumn{1}{c}{\bf T} &\multicolumn{1}{c}{\bf L} &\multicolumn{1}{c}{\bf B} \\
\hline \\
PixelCNN++, 1 stream & 3.07 & 3.10 & 3.05 \\
PixelCNN++, 2 stream & 2.97 & 2.98 & 2.93\\
Ours (1 stream, adversarial order)  & 2.93 & 2.98 & 3.05 \\
Ours (1 stream, 1 max context order)  & 2.77 & 2.83 & 2.89 \\
Ours (1 stream, 2 max context orders)  & \textbf{2.76} & \textbf{2.82} & \textbf{2.88}
\end{tabular}}
\end{center}
\vspace{-4mm}
\end{table}

On CIFAR10, we achieve 2.89 bpd test set likelihood when averaging the joint probability of 8 graphical models, each defined by an S-curve generation order. Our results outperform the state-of-the-art convolutional autoregressive model, PixelCNN++. We significantly outperform a 1 stream architectural variant of PixelCNN++ that has the same number of parameters as our model and uses a similar architecture, differing only in that it uses a single raster scan order. By introducing order-agnostic ensemble averaging to convolutional autoregressive models, we combined the best of fully-connected density estimators that average over orders, and the inductive biases of CNNs. These results could further improve with self-attention mechanisms and additional capacity, which have been observed to improve the performance of singe-order estimation, marking an opportunity for future research.

\begin{figure}[t]
	\centering
    \includegraphics[width=\linewidth]{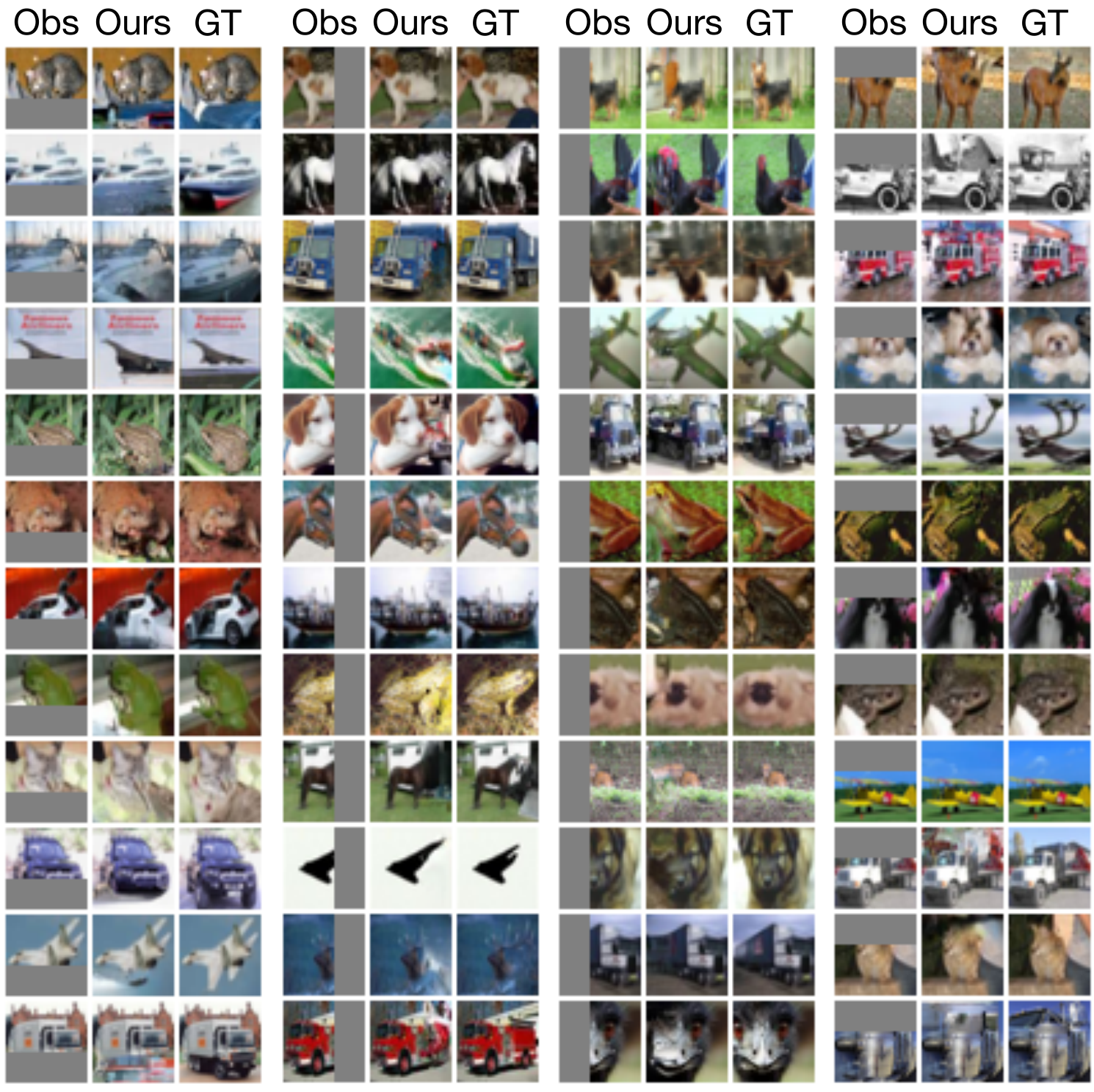}
    \vspace{-6mm}
	\caption{CIFAR10 image completions using our locally-masked convolutions with a specialized ordering.}
	\label{fig:cifar_inpainting}
	\vspace{-2mm}
\end{figure}

Our model is also scalable to high resolution distribution estimation. On the CelebA-HQ 256x256 dataset at 5-bit color depth, our model achieves 0.74 bpd with a single S-curve order, outperforming Glow \citep{kingma2018glow}, an exact likelihood normalizing flow. In comparison, the state-of-the-art model, SPN \citep{menick2018generating}, achieves 0.61 bpd by using self-attention and a specialized architecture for high resolutions.

\begin{figure*}[t]
	\centering
	\includegraphics[width=\linewidth]{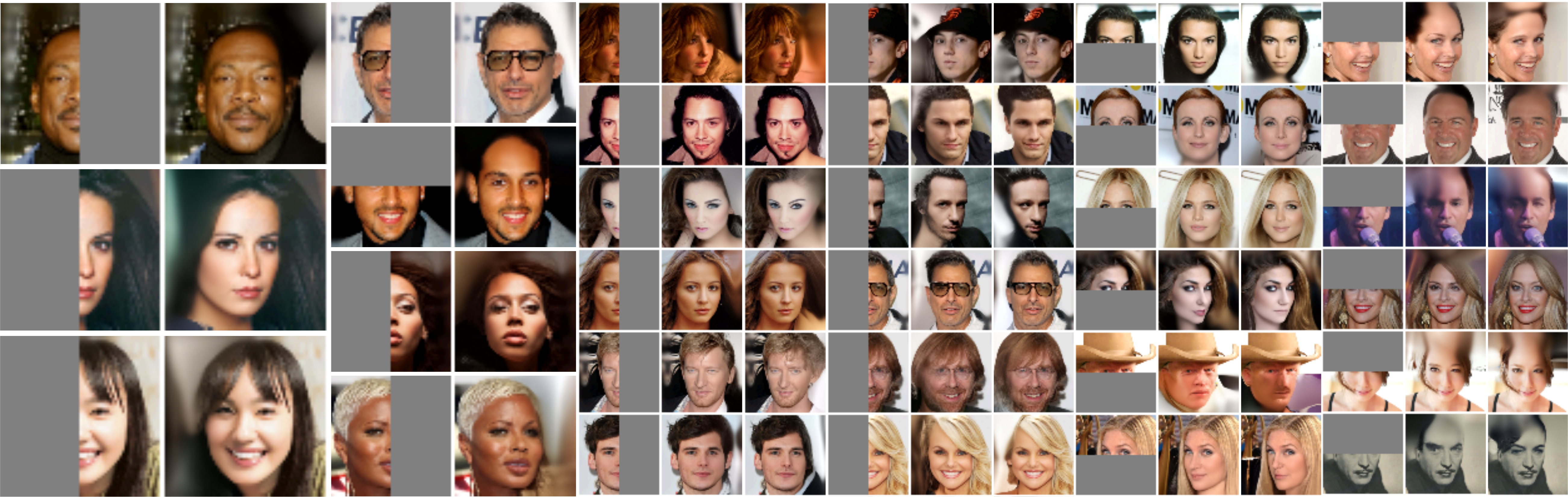}
	\vspace{-7mm}
	\caption{Completions of $64\times64$ px CelebA-HQ images at 5-bit color depth. Up to 2 samples are shown to the right of each half-obscured face provided to the model. Missing pixels are generated along an S-curve that first traverses the observed region. Additional samples and ground truth completions are provided in the appendix.}
	\label{fig:celebahq_inpainting_more}
\end{figure*}

\subsection{GENERALIZATION TO NOVEL ORDERS}

Ideally, an order-agnostic model would be able to generate images in orders that it has not been trained with.
To understand generalization to novel orders, we evaluate the test-set likelihood of a CIFAR10 model that achieves 2.93 bpd with a single S-curve order and 2.91 bpd with 8 S-curve orders under a raster scan decomposition. The model achieves 3.75 bpd with 1 raster scan order (28\% increase) and 3.67 bpd with 8 raster scan orders (26\% increase). While the novel order degrades compression rate, the model was trained with 8 fixed orders of the same S-curve type, which are fairly different from a raster scan.

To study generalization to more similar orders, we trained a model on Binarized MNIST with 7 S-curves for 120 epochs. On the test set, the model has 0.144 bpd using each train order. Testing with the held out (8th) S-curve, the model achieves 0.151 bpd, only 5\% higher.

\subsection{IMAGE COMPLETION}

To quantitatively assess whether control over generation order improves image completions, we measure the average conditional negative log likelihood of hidden regions of held-out test images on the MNIST and CIFAR10 datasets, measured in bits per dimension. We compute the NLL of the top half, left half, and bottom half of the image conditioned on the remainder of the image. The hidden region is set to zero in the model input, as well as hidden via masks used in each model.

Table~\ref{table:mnistbpd} shows average NLL on binary MNIST and CIFAR10. Top half inpainting is challenging for PixelCNN baselines that use a raster scan order, as model conditional $p_\mathbf{\theta}(x_i | x_{<i})$ does not condition on observed pixels that lie below $x_i$ in the image. Similarly, our architecture under an adversarial order, a single S-shaped curve from the top left to bottom left of the image, achieves $2.93$ bpd on CIFAR in the \textbf{T} setting. In contrast, using the same parameters, when we decomposes the joint favorably for maximum context with an S-curve generation order from the bottom left to the top left of the image, we achieve $2.77$ bpd. Averaging over two maximum context orders further improves log likelihood to $2.76$ bpd. A similar trend is observed for the other completion tasks, \textbf{L} and \textbf{B}.

\subsection{QUALITATIVE RESULTS}

Figure~\ref{fig:inpainting} shows completions of MNIST and CelebA-HQ 64$\times$64 images. PixelCNN++ produces MNIST digits that are inconsistent with the observed context. With a poor choice of order, our model only respects some attributes of the input image, but not overall facial structure. The model distributions over each missing pixel should condition on the entire observed region. This is accomplished when the missing region is generated last via a maximum context order. With this order, completions by our model are consistent with the given context.

Figures~\ref{fig:cifar_inpainting} and~\ref{fig:celebahq_inpainting_more} show completions of held-out CIFAR10 32$\times$32 and CelebA-HQ 64$\times$64 images for four different missing regions.
The masked input to the model (Obs), our sampled completion (Ours) and the ground truth image (GT) are shown. Missing image regions are generated in a maximum context order. While samples have some artifacts such as blurring due to long sequence lengths, images are globally coherent, with matching colors and object structure (CIFAR10) or facial structure (CelebA-HQ). Across datasets and image masks, our model effectively uses available context to generate coherent samples.

\section{RELATED WORK}
Autoregressive models are a popular choice to estimate the joint distribution of high-dimensional, multivariate data in deep learning. 
\cite{frey1998graphical} proposes logistic autoregressive Bayesian networks where each conditional is learned through logistic regression, capturing first-order dependencies between variables. While different orders had similar performance, averaging densities from 10 differently ordered models achieved small improvements in likelihood.
\cite{bengio2000modeling} extend this idea, using artificial neural networks to capture conditionals with some parameter sharing.
\cite{larochelle2011neural} propose the neural autoregressive distribution estimator (NADE) for binary and discrete data, reducing the complexity of density estimation from quadratic in the number of dimensions to linear. \cite{uria2013rnade} extend NADE to real-valued vectors (RNADE), expressing conditionals as mixture density networks. The autoregressive approach is desirable due to the lack of conditional independence assumptions, easy training via maximum likelihood, tractable density, and tractable, though sequential, forward sampling directly from the conditionals.

These works all use a single, arbitrary order per estimated model. However, it is possible to use the same parameters to define a family of differently ordered autoregressive Bayesian networks. \cite{uria2014deep} propose EoNADE, an ensemble of input-masked NADE models trained with an order-agnostic training procedure that achieve higher likelihoods when averaged and allows forward sampling of arbitrary regions. Each iteration, EoNADE chooses a random prefix of an ordering $\pi(1),\ldots,\pi(d)$, sample a training example $x$ and maximize the likelihood of $x_d$ under their model. ConvNADE~\citep{JMLR:v17:16-272} adapts EoNADE with a convolutional architecture and conditions the model on the input mask defining the order. 
Still, NADE, EoNADE and ConvNADE are serial: only a single conditional is trained at a time, and density estimation requires $D$ passes. \cite{germain2015made} propose an order-agnostic MADE that masks the weights of a fully connected autoencoder to estimate densities with a single forward pass by computing conditionals in parallel. While MADE supports multiple orders, it is limited by a fully-connected architecture. Our Locally Masked PixelCNN can be seen as a generalization of MADE that supports convolutional inductive bias.

Other deep autoregressive models use recurrent, convolutional or self-attention architectures.
In language modeling, autoregressive recurrent neural networks (RNNs) predict a distribution over the next token in a sequence conditioned on a recurrently updated representation of the previous words \citep{mikolov2010recurrent}. \cite{oord2016pixel} extend this idea to images, proposing a multi-dimensional, sequential PixelRNN for image generation and discrete distribution estimation, and a parallelizable PixelCNN. 
Subsequent works capture correlations between pixels in an image with convolutional architectures inspired by the PixelCNN \citep{van2016conditional, salimans2017pixelcnnpp, menick2018generating, reed2017parallel}, often improving the ability of the network to capture long-range dependencies. The PixelCNN family can generate entire high-fidelity images and, until recently, achieved state-of-the-art test set likelihood among tractable, likelihood-based generative models. PixelCNNs have also been used as a prior for latent variables~\citep{van2017neural}, and can be sampled in parallel using fixed-point methods~\citep{song2020nonlinear, wiggers2020predictive}. While convolutions process information locally in an image, self-attention mechanisms have been used to gain global receptive field \citep{chen2017pixelsnail, parmar2018image, child2019generating} for improved statistical performance.

Normalizing flows \citep{rezende2015variational} are parametric density estimators that give exact expressions for likelihood using the change-of-variables formula by transforming samples from a simple prior with learned, invertible functions. If tractable densities are not required, other families are possible. Implicit generative models such as GANs \citep{goodfellow2014generative} have been applied to high resolution image generation \citep{karras2018progressive} and inpainting \citep{pathak2016context}. Nonparametric approaches have also been successful for inpainting~\citep{efros1999texture,Hays:2007, Barnes:2009:PAR}. Partial convolutions~\citep{liu2018partialinpainting} improve CNN inpainting quality by rescaling filter responses that access missing pixels, but are not causal unlike \ours{}. Latent-variable models like the VAE \citep{kingma2013auto, rezende2014stochastic} jointly learn a generative model for data $x$ given latent $z$ and an approximation for the posterior over $z$. Other latent-variable models are based on Markov chains \citep{bengio2014deep, sohl2015deep, nijkamp2019learning}.

\section{CONCLUSION}
In this work, we proposed an efficient, scalable and easy to implement approach for supporting arbitrary autoregressive orderings within convolutional networks. To do so, we propose \textit{locally masked convolutions} that allow arbitrary orderings by masking features at each layer while simultaneously sharing filter weights. This formulation can be efficiently implemented purely via matrix multiplication. 
Our work is a synthesis of prior lines of inquiry in autoregressive models. Locally Masked PixelCNNs support parallel estimation, convolutional inductive biases, and control over order, all with one simple layer. Foundational work in this area each supported some of these, but with incompatible architectures.
As an additional benefit, arbitrary orderings allow image completion with diverse regions.
We achieve globally coherent image completions by choosing a favorable order at test time, without specifically training the model to inpaint.

\subsubsection*{Acknowledgements}
We thank Paras Jain, Nilesh Tripuraneni, Joseph Gonzalez and Jonathan Ho for helpful discussions, and reviewers for helpful suggestions. This research is supported in part by the NSF GRFP under grant number DGE-1752814, Berkeley Deep Drive and the Open Philanthropy Project. Any opinions, findings, and conclusions or recommendations expressed in this material are those of the authors and do not necessarily reflect the views of the NSF.

\clearpage
\nocite{gilbert}
\subsubsection*{References}
\begingroup
\small
\renewcommand{\section}[2]{} 
\setlength{\bibsep}{1pt}
\bibliography{main}
\endgroup

\clearpage
\appendix
\section{APPENDIX}
\vspace{-2mm}
\subsection{ORDER VISUALIZATION}
\begin{figure}[bh]
    \centering
    \includegraphics[width=0.775\linewidth]{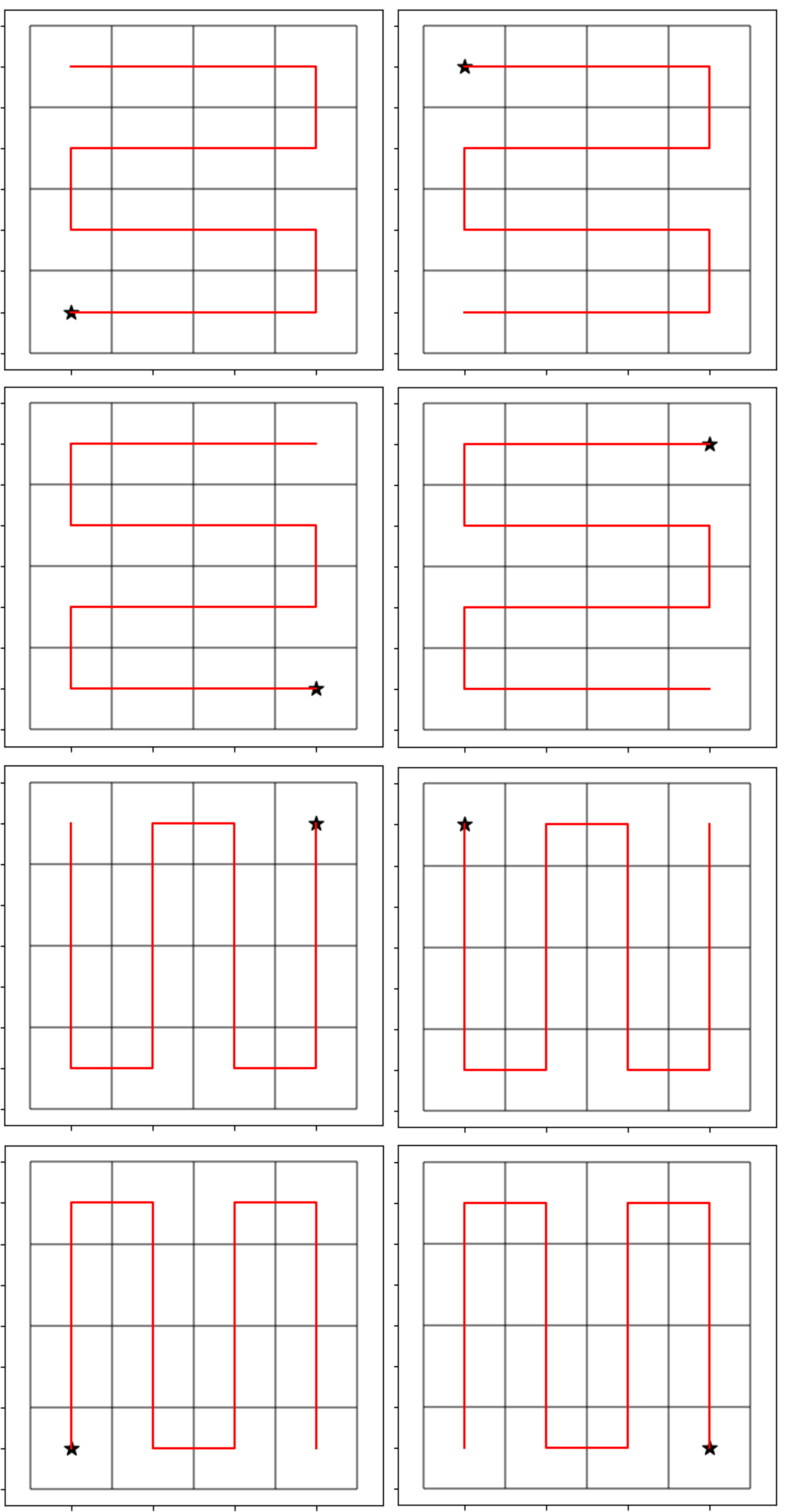}
    \caption{Eight variants of the S-curve generation order.}
    \label{fig:variants}
\end{figure}

\vspace{-2mm}
\label{sec:mask_viz}
Figure~\ref{fig:orders_viz} shows three image generation orders and corresponding local masks used by the first \ours{} layer in the autoregressive generator. On the left, we show the raster scan, S-curve and Hilbert curve orders over the pixels of a small 8$\times$8 image. On the right, we show the corresponding, local 3$\times$3 binary masks applied to image patches in the first layer. Masks applied to zero-pad pixels are colored green as their value is arbitrary. The center pixel in each image patch is masked out (set to 0), so that the network cannot include ground truth information in the representation of its context. The raster scan masks are the same for all image patches, so weights can be masked rather than image patches. However, other orders require diverse masks to respect the autoregressive property of the model. Figure~\ref{fig:variants} shows the 8 variants of the S-curve generation order used for order-agnostic training.

\subsection{MASK CONDITIONING}
\vspace{-2mm}
\cite{JMLR:v17:16-272} propose a convolutional neural autoregressive distribution estimator (ConvNADE) that can be trained with different masks on the input image. ConvNADE concatenates the mask with the image, allowing the model to distinguish between a zero-valued pixel and a zero-valued mask. Locally masked convolutions can also condition upon the mask in each layer. Algorithm~\ref{alg:lmconv_mc} is an adaptation of Algorithm~\ref{alg:lmconv} that supports mask conditioning, with modifications shown in green. Algorithm~\ref{alg:lmconv_mc} applies a learned weight matrix $\mathcal{W}_\mathcal{M}$ to the first $C_\text{in}$ rows of the mask matrix as the mask is repeated $k_1*k_2$ times by Algorithm~\ref{alg:makemasks}. Equivalently, the mask $\mathcal{M}_{1:C_\text{in}}$ can be concatenated with $X$ after masking.

We evaluate mask conditioning on the Binarized MNIST dataset with 8 S-curve orders. After training for 60 epochs (not converged for the purposes of comparison), the model without mask conditioning achieves a test NLL of 77.85 nats, while the mask conditioned model achieves a comparable test NLL of 77.94 nats. However, mask conditioning could improve generalization to novel orders.

\begin{algorithm}[t]
  \caption{\textsc{LMConv} with {\color{ForestGreen} mask conditioning}}
  \label{alg:lmconv_mc}
\begin{algorithmic}[1]
  \STATE {\bfseries Input:} image $x$, weights $\mathcal{W}_X, {\color{ForestGreen} \mathcal{W}_\mathcal{M} }$, bias $b$, generation order $\pi$. $x$ is ${B \times C_\text{in} \times H \times W}$, $\mathcal{W}_X$ is ${C_\text{out} \times C_\text{in} * k_1 * k_2}$, and {\color{ForestGreen} $\mathcal{W}_\mathcal{M}$ is ${C_\text{out} \times k_1 * k_2}$}.
  \STATE Create mask matrix $\mathcal{M}$ with Algorithm~\ref{alg:makemasks}
  \STATE Extract patches: $X = \texttt{im2col}(\texttt{pad}(x), k_1, k_2)$
  \STATE Mask patches: $X = \mathcal{M} \odot X$
  \STATE Perform convolution: $Y = \mathcal{W}_XX {\color{ForestGreen} + \mathcal{W}_\mathcal{M} \mathcal{M}_{1:C_\text{in}} } + b$
  \STATE Assemble patches: $y = \texttt{col2im}(Y)$
  \STATE {\bfseries return} $y$
\end{algorithmic}
\end{algorithm}

\subsection{EXPERIMENTAL SETUP}
\label{sec:experimental_setup}
\vspace{-2mm}

We tune hyperparameters such as the learning rate and batch size as well as the network architecture (Section~\ref{sec:architecture}) on the Grayscale MNIST dataset, and train models with the exact same architecture and hyperparameters on Binarized MNIST, CIFAR and CelebA-HQ. We used a batch size of 32 images, learning rate $2\times10^{-4}$, and gradient clipping to norm $2\times 10^6$. The exception is that we use batch size 5 on CelebA-HQ to save memory and 2-way softmax output instead of logistics for binary data. CelebA-HQ~\citep{karras2018progressive} contains 30,000 $256\times256$ 8-bit color celebrity photos. For experiments, we use the same CelebA-HQ data splits as Glow~\citep{kingma2018glow}, with 27,000 training images and 3,000 validation images at reduced 5-bit color depth.

We trained the 1 stream baseline and our model for about the same number of epochs. Longer training improves performance, perhaps because order-agnostic training and dropout regularize, so epoch count was determined by time limitations. Most models are trained with 4 V100 or Quadro RTX 6000 GPUs. We train our CIFAR10 model for 2.6M steps (1644 epochs) with order-agnostic training over 8 precomputed S-curve variants, then average model parameters from the last 45 epochs of training. Early in our experimental process, we compared Hilbert curve generation orders against the S-curve, visualized for small images in Figure~\ref{fig:orders_viz}, but did not see improved results.

For qualitative results, we train the 184\textsc{M} parameter ${64\times64}$ CelebA-HQ model for 375\textsc{K} iterations at batch size 32. Inspired by Progressive GAN~\citep{karras2018progressive}, we train the model at a reduced $32\times32$ resolution for the first 242\textsc{K} iterations. As the architecture is fully convolutional, it is straightforward to increase image resolution during training.

\subsection{ADDITIONAL SAMPLES}

\begin{figure}
    \centering
    \includegraphics[width=\linewidth]{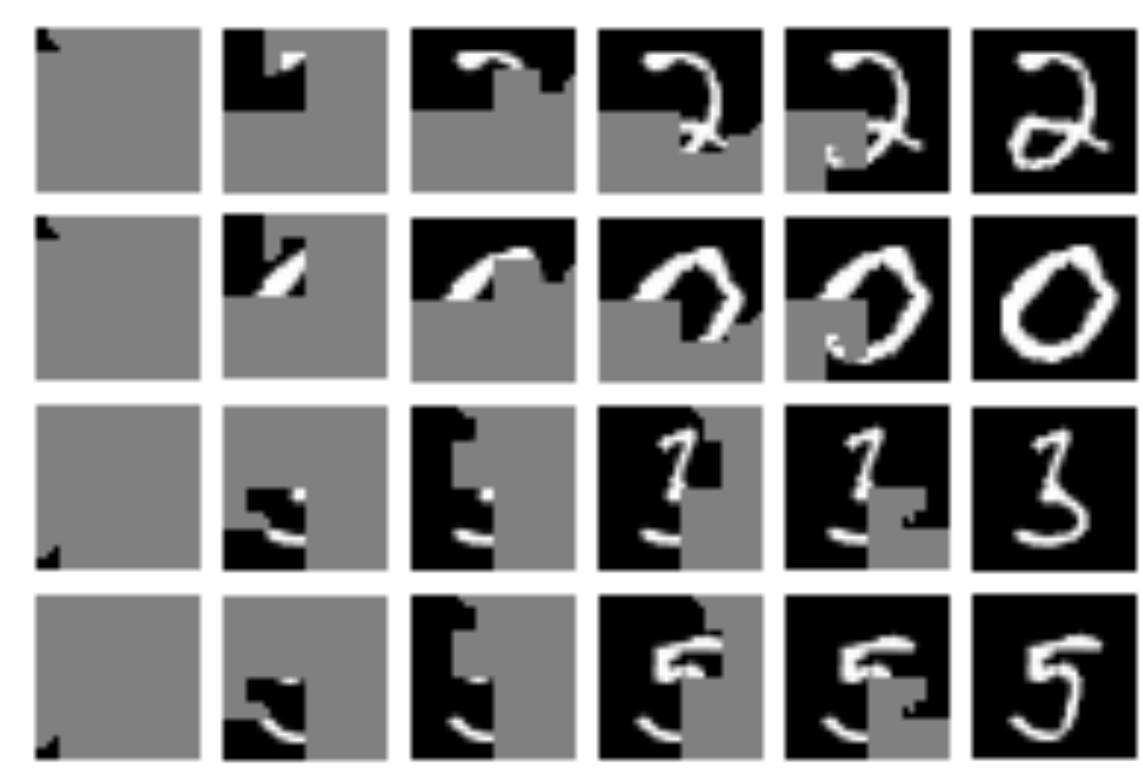}
    \vspace{-6mm}
    \caption{Unconditionally generating MNIST digits with two Hilbert curve orders, starting at the top or bottom left.}
    \label{fig:mnist_generation}
\end{figure}

Figure~\ref{fig:mnist_generation} shows intermediate states of the forward sampling process for unconditional generation of grayscale MNIST digits. We samples pixels along a Hilbert space-filling curve. As Hilbert curves are defined recursively for power-of-two sized grids, we use a generalization of the Hilbert curve~\citep{gilbert} for $28\times28$ image generation. Our Locally Masked PixelCNN is optimized via order-agnostic training with eight variants of the order. Two variants are used for sampling digits in Fig.~\ref{fig:mnist_generation}. The top two digits are sampled beginning at the top left of the image, and the bottom two digits are sampled beginning at the bottom left of the image. Images are shown at intervals of roughly 156 sampling steps. With the same parameters, the model is able to unconditionally generate plausible digits in multiple orders.

\begin{figure}[t!]
	\centering
	\includegraphics[width=\linewidth]{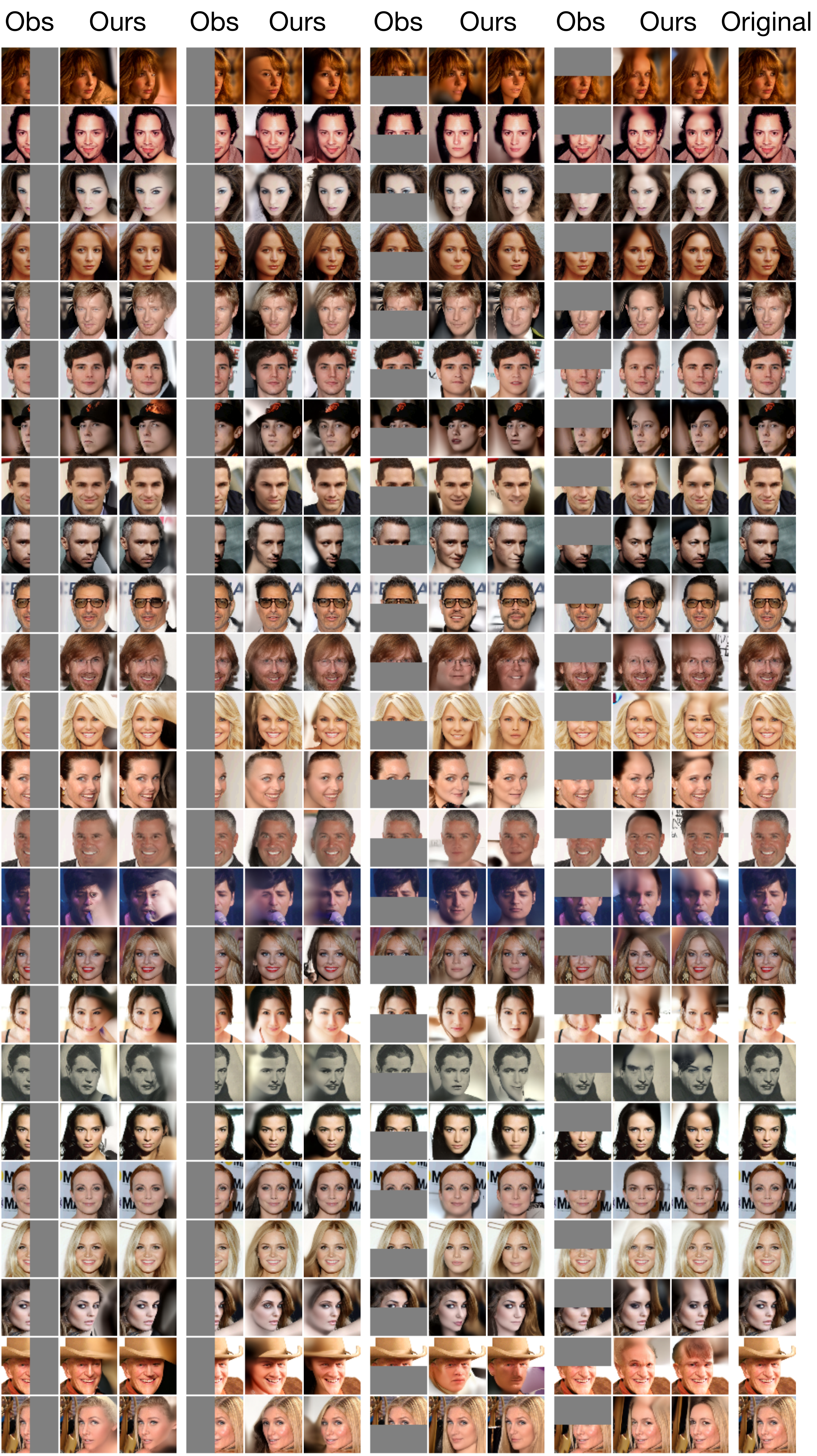}
	\vspace{-6mm}
	\caption{Uncurated CelebA-HQ 64x64 completions.}
	\label{fig:celebahq_inpainting_extra}
\end{figure}

Figure~\ref{fig:celebahq_inpainting_extra} shows uncurated image completions using the large CelebA-HQ model. Initial network input is shown to the left of two image completions sampled from our Locally Masked PixelCNN with an S-curve variant that generates missing pixels last. The input images are taken from the validation set. The rightmost column contains the original image, \ie{} the ground truth image completion. Two samples with the same context vary due to the stochasticity of the decoding process, \eg{} varying in terms of hairstyle, facial hair, attire and expression.

\subsection{IMPLEMENTATION}

Locally Masked Convolutions are simple to implement using the basic linear algebra subprograms exposed in machine learning frameworks, including matrix multiplication. It also requires an implementation of the im2col operation. We provide an abbreviated Python code sample implementing \ours{} using the PyTorch library in Figure~\ref{fig:implementation}. The full source including gradient computation, parameter initialization and mask conditioning is available at
\href{https://ajayjain.github.io/lmconv}{https://ajayjain.github.io/lmconv}.

\begin{figure*}
\begin{lstlisting}[language=Python]
import math

import torch 
import torch.nn as nn
import torch.nn.functional as F

class _locally_masked_conv2d(torch.autograd.Function):
    @staticmethod
    def forward(ctx, x, mask, weight, bias=None, dilation=1, padding=1):
        # Save values for backward pass
        ctx.save_for_backward(x, mask, weight)
        ctx.dilation, ctx.padding = dilation, padding
        ctx.H, ctx.W = x.size(2), x.size(3)
        ctx.output_shape = (x.shape[2], x.shape[3])
        out_channels, in_channels, k1, k2 = weight.shape

        # Step 1: Unfold (im2col)
        x = F.unfold(x, (k1, k2), dilation=dilation, padding=padding)

        # Step 2: Mask x. Avoid repeating mask in_channels times by reshaping x
        x_channels_batched = x.view(x.size(0) * in_channels,
            x.size(1) // in_channels, x.size(2))
        x = torch.mul(x_channels_batched, mask).view(x.shape)

        # Step 3: Perform convolution via matrix multiplication and addition
        weight_matrix = weight.view(out_channels, -1)
        x = weight_matrix.matmul(x)
        if bias is not None:
            x = x + bias.unsqueeze(0).unsqueeze(2)

        # Step 4: Restore shape
        return x.view(x.size(0), x.size(1), *ctx.output_shape)

    @staticmethod
    def backward(ctx, grad_output):
        x, mask, weight, mask_weight = ctx.saved_tensors
        ...
        if ctx.needs_input_grad[2]:
            # Recompute unfold and masking to save memory
            x_ = F.unfold(x, (k1, k2), dilation=ctx.dilation, padding=ctx.padding)
            ...
        ...

class locally_masked_conv2d(nn.Module):
    def __init__(self, in_channels, out_channels, kernel_size, dilation, bias):
        super(locally_masked_conv2d, self).__init__()
        ...
        self.weight = nn.Parameter(torch.Tensor(out_channels, in_channels, *kernel_size))
        self.bias = nn.Parameter(torch.Tensor(out_channels)) if bias else None
        self.reset_parameters()

    def reset_parameters(self):
        ...

    def forward(self, x, mask):
        return _locally_masked_conv2d.apply(x, mask, self.weight,
            self.bias, self.dilation, self.padding)
\end{lstlisting}
\caption{A memory-efficient PyTorch v1.5.1 implementation of \ours{}. Gradient calculation is omitted for brevity. See \url{https://ajayjain.github.io/lmconv} for the full implementation and training code.}
\label{fig:implementation}
\end{figure*}

\end{document}